\documentclass[letterpaper, 10 pt, conference]{ieeeconf}

\IEEEoverridecommandlockouts                              % This command is only needed if 
\usepackage{amsmath,amsfonts}
\usepackage{algorithmic}
\usepackage{algorithm}
\usepackage{array}
\usepackage[caption=false,font=normalsize,labelfont=sf,textfont=sf]{subfig}
\usepackage{textcomp}
\usepackage{stfloats}
\usepackage{url}
\usepackage{verbatim}
\usepackage{graphicx}
\usepackage{cite}
\usepackage{xcolor}
\usepackage{booktabs}
\usepackage{multirow}
\usepackage{ulem}
\usepackage{gensymb}
\usepackage{float}
\usepackage[pdfencoding=auto,psdextra,breaklinks,colorlinks]{hyperref}
\usepackage{subcaption}
\newcommand{\blue}[1]{#1}
\newcommand{\red}[1]{#1}
\newcommand{\mgn}[1]{#1}
\newcommand{\brn}[1]{#1}

\begin{document}

\title{Extremum Seeking Controlled Wiggling for Tactile Insertion}

%\iffalse
\author{Levi Burner$^{1*}$, Pavan Mantripragada$^{2*}$, Gabriele M. Caddeo$^{3*}$, Lorenzo Natale$^{3}$,\\ Cornelia Ferm\"uller$^{4}$, Yiannis Aloimonos$^{2,4}$
        % <-this % stops a space
\thanks{$^{*}$ Equal Contribution}
\thanks{$^{1}$ Corresponding author. Department of Electrical and Computer Engineering, University of Maryland, College Park,
        {\tt\small lburner@umd.edu}}
\thanks{$^{2}$ Department of Computer Science, University of Maryland, College Park,
        {\tt\small mppavan@umd.edu}}
\thanks{{$^{3}$ Istituto Italiano di Tecnologia, Via San Quirico, 19 D, Genova, Italy.
        {\tt\small \{gabriele.caddeo, lorenzo.natale\}@iit.it}}}
\thanks{$^{4}$ University of Maryland Institute for Advanced Computer Studies, University of Maryland, College Park
        {\tt\small \{fer, jyaloimo\}@umiacs.edu }}
\thanks{Code and experimental data are available at \url{prg.cs.umd.edu/ESTac}}
}
%\fi
%\author{Anonymous Authors$^{*}$% <-this % stops a space
% \thanks{$^{*}$ Code anonymized for review is available at: \url{https://anonymous.4open.science/r/WigglingInsertion-Anonymous-0F22}}% <-this % stops a space
% \thanks{All code and data will be released upon the papers acceptance.}% <-this % stops a space
% }


%. Preprint Version. Accepted Month, Year}
%{FirstAuthorSurname \MakeLowercase{\textit{et al.}}: ShortTitle} 


\maketitle

\begin{abstract}
When humans perform \mgn{complex} insertion tasks such as \mgn{pushing} a cup into a cupboard, routing a cable, or \mgn{putting} a key in a lock, they wiggle the object and \mgn{adapt} the process through tactile feedback. \mgn{A similar robotic approach has not been developed}. \mgn{We} study an extremum seeking control law that wiggles end effector pose to maximize insertion depth while minimizing strain measured by a GelSight Mini sensor.
\mgn{Evaluation is conducted} on \blue{four \mgn{keys featuring complex geometry} and five \mgn{assembly} tasks \mgn{featuring basic geometry}.}

\blue{On keys, the algorithm achieves 71\% success rate over 120 trials with 6-DOF perturbations, 84\% over 240 trials with 1-DOF perturbations, and 75\% \mgn{over 40 trials} initialized with vision.}
\mgn{It} \blue{significantly outperforms a baseline optimizer, CMA-ES, \mgn{that replaces wiggling with random sampling}}\mgn{. When tested on a state-of-the-art assembly benchmark featuring basic geometry, it achieves 98\% over 50 vision-initialized trials. The benchmark's most similar baseline, which was trained on the objects, achieved 86\%.}
\mgn{These results, \blue{realized without contact modeling or learning,} show that closed loop wiggling based on tactile feedback is a robust paradigm for robotic insertion.}
\end{abstract}

% \begin{IEEEkeywords}
% Dexterous Manipulation, Perception for Grasping and Manipulation, Assembly, Contact Modeling
% \end{IEEEkeywords}

\section{Introduction}
Imagine inserting a cup into a crowded cupboard. There are many objects in the way of where you want to put the cup, but somehow as the cup is inserted, they are pushed aside and the cup is placed in the appropriate location. A similar process occurs when routing a cable through a hole or inserting a key into a lock.
%We call such insertion problems crowded, through, and direct, respectively, and note that
Humans can accomplish all of them from a young age but robots still struggle. Further, general-purpose robots will need to solve such problems frequently, \blue{from peg-in-hole tasks featuring simple geometry, to more complex ones like key insertion. Key insertion offers unique challenges compared to the traditionally studied peg-in-hole because locks contain numerous internal contact surfaces which are difficult to model.}

\mgn{During such insertion tasks, humans often wiggle the object to be inserted while using tactile feedback from the fingers to adjust the process and maximize insertion depth.}
Thus, it is interesting to study generalizable insertion algorithms, inspired by human behavior, that use strain-like tactile feedback and control laws based on wiggling motions.

\blue{State-of-the-art robotic insertion methods typically focus on policies tailored to specific objects \cite{kim2022activeextrinsiccontactsensing, tang2024automate} even when incorporating tactile feedback \cite{kim2022activeextrinsiccontactsensing, kamijo2023tactilebasedactiveinferenceforcecontrolled}. Approaches that aim for object generalization often rely on strong assumptions about initial conditions \cite{yan_2021learning}. Recently, learning-based techniques have gained traction, but these are frequently trained and evaluated on the same set of objects \cite{tang2024automate, wu2024tacdiffusion}.
In contrast, we demonstrate that real-time tactile feedback can successfully guide a learning-free wiggling-based insertion process that generalizes to new objects \mgn{without requiring} detailed contact-modeling, \mgn{prior knowledge of} 3D geometry, \mgn{or} strong assumptions on the starting pose.}

\begin{figure}[t]
  \centering
  \includegraphics[width=1.0\linewidth]{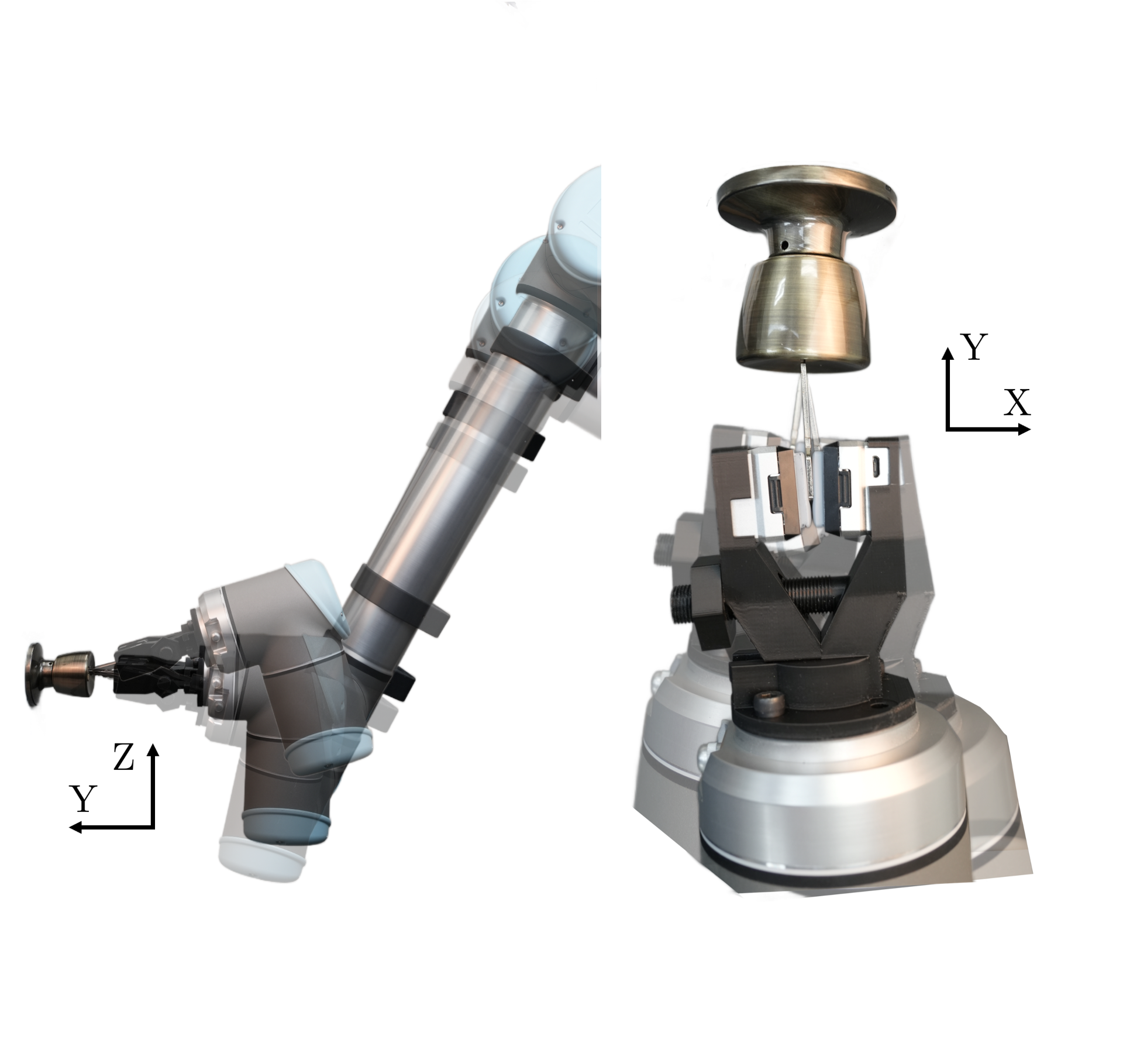}
  \caption{By wiggling the 6 degree of freedom pose of an object grasped between two GelSight Mini tactile sensors and observing a strain-like quantity through the optical flow in the GelSight cameras, an extremum seeking control law performs insertion. All parameters are sinusoidally modulated simultaneously but at different frequencies, allowing for the estimation of a direction that minimizes strain and maximizes insertion depth along the Y axis.}
  \label{fig:overview}
\end{figure}

\begin{figure*}[htbp]
  \centering
  \vspace*{0.1in}
  \includegraphics[width=1.0\linewidth]{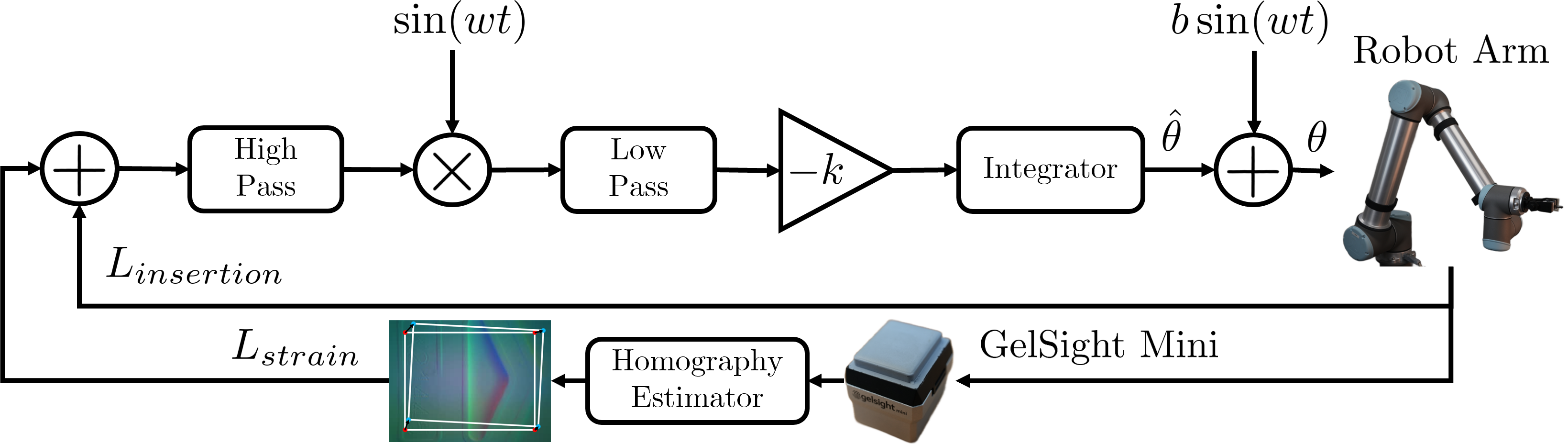}
  \caption{The extremum seeking controlled pipeline for wiggling-based tactile insertion. The instantaneous parameters $\theta$ control the pose of the tip of an object through a UR10 robot arm. The strain that the object exerts on the GelSight Mini's gel pad is observed via a displacement of the corners of a tracked patch in the sensor image feed, $L_{strain}$. The objective to be minimized is the sum of $L_{strain}$ plus $L_{insertion}$, where $L_{insertion}$ represents the depth of insertion into the lock. The extremum seeking control seeks to minimize the objective by adjusting the parameter estimates $\hat{\theta}$. As is standard in Extremum Seeking Control, $\theta$ is a modulated version of $\hat{\theta}$ with each parameter modulated at a different frequency. The high pass filter removes the DC component from the objective signal, demodulation determines the slope of the objective's gradient, and the low pass filter averages the feedback signal with greater high-frequency attenuation than the integrator.}
  \label{fig:system}
\end{figure*}

\blue{We consider a} compliant gripper, as shown in Figure \ref{fig:overview}, \blue{which grasps a key (in general, a peg) between}
two GelSight Mini tactile sensors \cite{yuan_2017}. The pads experience strain when attempting to insert the key into a lock, which \blue{is} observed through internal cameras that point toward the back of the gel.
\blue{A} control strategy based on wiggling \blue{is} realized through  Extremum Seeking Control, a 100-year-old method for \blue{online} optimization of objective functions \cite{leblanc1922electrification}. Due to its model-free formulation, it finds application in systems that are difficult to model, such as wind turbines and solar panels (which must handle changing weather), control of particle beams, and aircraft that experience flow instability \cite{SCHEINKER2024111481}. Despite its generality, the method has received little attention in the robotics community.

When applied to tactile insertion, the extremum seeking control wiggles the 6 degree of freedom pose of the end-effector to estimate a descent direction that minimizes \blue{an} objective \blue{depending on} tactile feedback and end-effector pose. A block diagram is given in Figure \ref{fig:system}. \blue{The method uses contact to correct pose, and so must be initialized near the opening, though not further than existing methods when \mgn{normalized by} opening size. We validate this assumption by conducting trials initialized with vision.}

A list of our contributions follows:
\begin{itemize}
\item A \blue{method for insertion, inspired by human behavior,} that uses \blue{tactile-feedback} to guide a wiggling process via Extremum Seeking Control \blue{that generalizes to all tested objects without per-object tuning.}
\item A method based on optical flow to measure strain-like forces on a GelSight Mini pad when grasping planar and rigid objects, such as keys.
\item Detailed evaluation
on key-in-lock insertion on four locks
amounting to 360 trials while significantly outperforming a baseline based on CMA-ES \cite{auger2012tutorial} \mgn{that randomly samples instead of wiggling}.
\item \blue{Vision-initialized experiments on key insertion and a robot assembly benchmark \cite{tang2024automate} (unseen until test time) that validate the method's assumptions and \mgn{demonstrate remarkable performance on the assembly benchmark.}}
\end{itemize}

\section{RELATED WORK}

Peg-in-hole problems are often considered in the field of robotics assembly. \mgn{Common techniques} have limitations \blue{when compared to our formulation based on closed loop wiggling} including: \red{requiring a contact model, the 3D model of the peg, or being restricted to specific object shapes or flat surfaces \cite{Park_2017,Park_2020b, Gibbons_2023, Nottensteiner_2020, kim2022activeextrinsiccontactsensing, dong2019tactile, kamijo2023tactilebasedactiveinferenceforcecontrolled, li2014localization, Fuchioka_2024, chen2025robustpeginholeassemblyuncertainties}. Others consider only small translation errors compared to the peg size \cite{kim2022activeextrinsiccontactsensing, dong2021tactilerlinsertiongeneralizationobjects, Fuchioka_2024}, or prior knowledge of the hole position \cite{li2014localization}.} While some works also use wiggling, they do so with feedforward motions, that is, without realtime adaptation based on feedback \cite{Park_2017, Park_2020b, Gubbi_2020, wu20241khz, wu2024tacdiffusion}. \blue{Often, methods focus on contact modeling for specific object and sensor configurations \cite{kim2022activeextrinsiccontactsensing, Gibbons_2023}. Our method does not use a contact model.}

Some works focus on problems adjacent to peg-in-hole including estimating the location of the grasped object within the grasp \cite{she2020cablemanipulationtactilereactivegripper}, vision based feedback \cite{visual-based-automation, relative-pose-estimation}, the role of force or impedance feedback \cite{Lee_2022gaussian, kamijo2023tactilebasedactiveinferenceforcecontrolled, Nottensteiner_2020, fuzzy, wu20241khz, wu2024tacdiffusion}, learning based approaches \cite{dong2021tactilerlinsertiongeneralizationobjects, lin2024generalizetouchingtactileensemble, Fuchioka_2024, tang2024automate, guo2025srsaskillretrievaladaptation, marougkas2025integratingmodelbasedcontrolrl, li2025easyinsertdataefficientgeneralizableinsertion}, and \blue{the manipulator's mechanical design \cite{fukaya2024fouraxisadaptivefingershand}. Our proposed approach is complementary to these ideas.}

\blue{GelSight sensors were used as source of tactile feedback in several of these studies \cite{kim2022activeextrinsiccontactsensing, kamijo2023tactilebasedactiveinferenceforcecontrolled, dong2021tactilerlinsertiongeneralizationobjects, she2020cablemanipulationtactilereactivegripper, li2014localization, dong2019tactile}}. One GelSight work also considers estimating a high-dimensional, dense optical flow on the gel pad surface during peg-in-hole insertion \cite{Wang_2024flow}, in contrast our method is low dimensional and can track grasped object pose over long time periods.

\blue{Key insertion} has received little attention. In \cite{yan_2021learning}, the problem of inserting a key into a bicycle lock is considered and solved using a bimanual approach \blue{while considering rotation uncertainty along a single axis and perfect translation estimates.} In \cite{chhatpar_2005}, the problem of \blue{localizing} a keyhole is tackled using a \blue{pre-determined} map of key-to-lock contact configurations. \mgn{A few recent works have considered key-in-lock insertion as part of a suite of test objects \cite{wu20241khz, wu2024tacdiffusion, li2025easyinsertdataefficientgeneralizableinsertion, zhao2025touchbeginsvisionends}. While \cite{wu20241khz, wu2024tacdiffusion} incorporate a wiggling strategy, it is open-loop.
Unlike these works, our approach considers closed-loop wiggling and shows that this paradigm can insert keys and objects, under 6D pose uncertainty, without sophisticated modeling, \mgn{learning, or per-object tuning.}}

While Extremum Seeking Control has found extensive application in traditional engineering fields \cite{SCHEINKER2024111481}, to our knowledge, it has not been used for robotic insertion. The closest work is an extremum seeking seeking control method for grasping with a gripper \cite{calli_2018}.

\blue{In summary,} \mgn{by developing a closed-loop wiggling process, we realize something} \red{unlike existing works\mgn{. That is,} a model-free approach to solve the tactile insertion task, making no specific assumptions about the object's shape, contact model, or restricting the \blue{uncertainty of the holes position to particular \mgn{degrees of freedom.}}}

\section{METHODS}

\blue{Our} algorithm \blue{that realizes closed-loop wiggling based on tactile feedback} has three parts. First, we describe the tactile feedback using strain measurements taken by a GelSight Mini tactile sensor and computed using a Lucas-Kanade style homography tracker. Next, we detail the objective function minimized by the extremum seeking control. Finally, we describe the extremum seeking control law, which sinusoidally perturbs the 6 degree of freedom pose of the end-effector to estimate a descent direction and minimize the objective function. A block diagram is given in Figure \ref{fig:system}.

\subsection{Tactile Strain Measurement}

\begin{figure}[t]
  \centering
  \vspace*{0.1in}
  \includegraphics[width=0.8\linewidth]{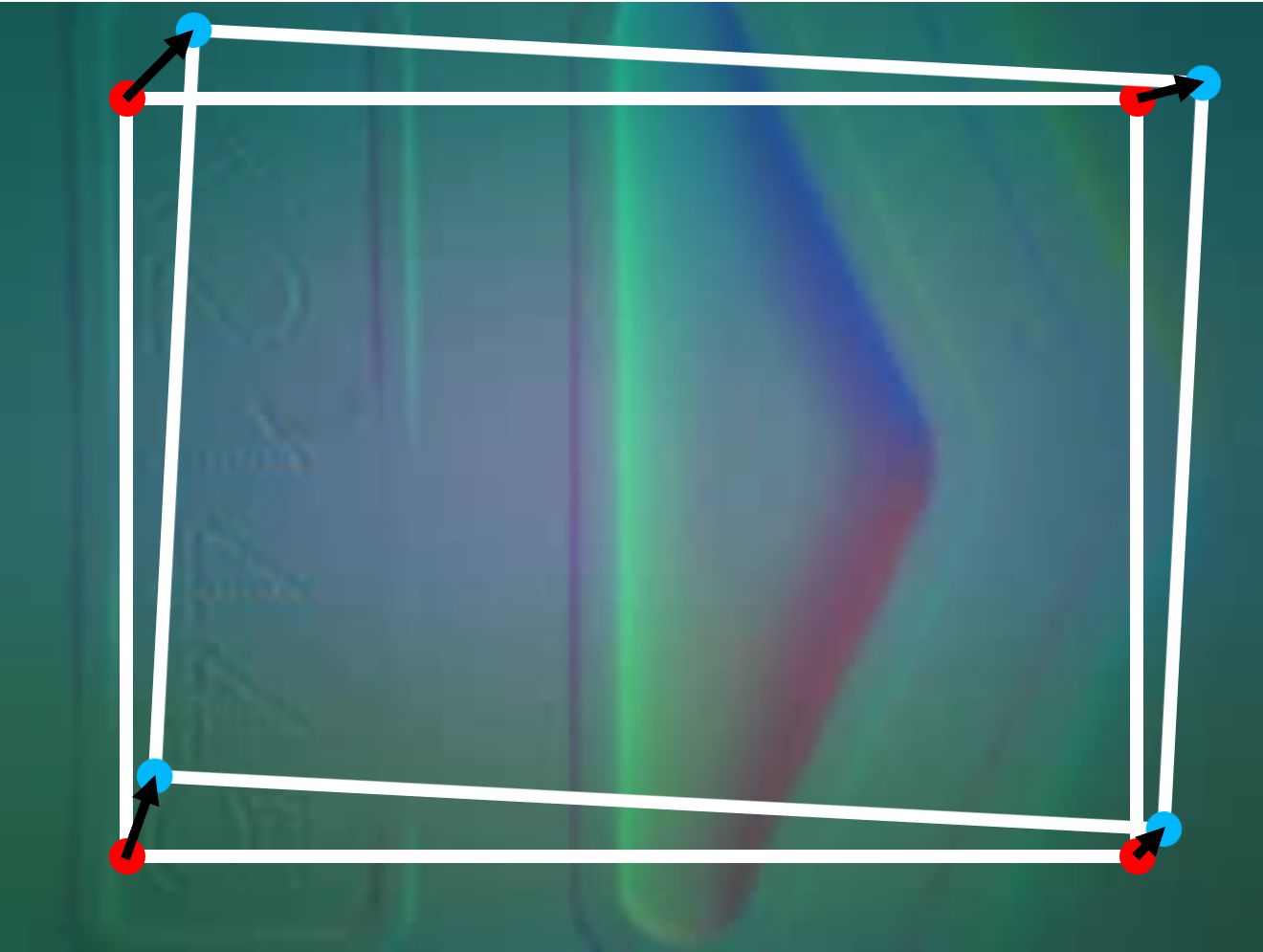}
  \caption{A strain-like measurement is measured directly from the images returned by the GelSight Mini sensor. Each incoming frame is iteratively registered with a Lucas-Kanade style homography estimator to the first frame. The tracked patch has 10\% margins with respect to the full frame and is exemplified by the area within the white box with red corners. The Euclidian norm of the corner displacements in pixels from their original location is used as the strain-like quantity $L_{strain}$.}
  \label{fig:gelsighttracking}
\end{figure}

The gripper configuration is shown in Figure \ref{fig:overview}. Two GelSight Mini tactile sensors are pressed against the left and right side of the key head. One tactile sensor is off and functions as a soft barrier. The second sensor is on, and the 2D imagery captured from the internal camera is exemplified in Figure \ref{fig:gelsighttracking}. Because the contact pads are compliant, when pressure is applied to the tip of the key, the key moves with respect to the gel pads and warps the 2D imagery. Further, because the head of key is flat, this warp is described as an 8-parameter homography. Thus, a Lucas-Kanade style homography estimator is used to continuously warp back the current image to the first frame. The optimization is warm-started at each iteration using the previously estimated homography. The implementation is based on \cite{baker2004lucas} and uses the 4 corner parameterization suggested in \cite{Baker_2006_9407}. Before passing through the tracker, frames are resized to 320$\times$240 resolution. The tracked patch is also initialized to the entire frame with 10\%  margins, as illustrated in Figure \ref{fig:gelsighttracking}.

The displacement of the four corners of the tracked patch from is used as the strain metric. That is

\begin{equation}
L_{strain}(t) = \sqrt{\sum_{i=1}^4 \left\|p_i(t) - p_i(t_0)\right\|^2},
\end{equation}
where $p_i(t)$ is the pixel location of the $i$'th corner at time $t$ as estimated by the homography tracker and $t_0$ is the time of the first frame. To avoid the effects of noise when pressure is low $L_{strain} < 3$ pixels is reported as $0$. The strain feedback is computed at 10-16 Hz. The update rate is limited by a combination of OpenCV's Python V4L2 camera interface, USB 2.0 bandwidth limits, and GelSight Mini's hardware requirement that images be retrieved at the high resolution of 3280$\times$2464 in the Motion-JPEG format.

\subsection{Control Objective}

To encourage the extremum seeking control to guide the key into the lock, an additional loss term $L_{insertion}$ is defined

\begin{equation}
L_{insertion} = |Y - (Y(t_0) - d)|,
\end{equation}
where $Y$ is the position of the tip of the key in meters along the axis to be inserted into the lock and $d$ is the depth of the keyhole.
If $L_{insertion} < 0.5$ millimeters, the algorithm is stopped and insertion is considered a success.

The final objective function is then
\begin{equation}
L = L_{insertion} + \lambda L_{strain},
\end{equation}
where $\lambda = 0.0005$.

\subsection{Extremum Seeking Control}

Classical extremum seeking control as described in \cite{SCHEINKER2024111481} is applied directly. A block diagram is given in Figure \ref{fig:system}. The estimated parameters $\hat{\theta}$ are the 6 degree-of-freedom pose of the tip of the key. That is $\hat{\theta} = [X, Y, Z, \alpha, \beta, \gamma]^T$ where $\alpha$, $\beta$, $\gamma$ are $x-y-z$ intrinsic Euler angles corresponding to the orientation of the tip of the key.

As is standard in extremum seeking control, the applied parameters $\theta(t)$ are modulated according to
\begin{equation}
\theta(t) = \hat{\theta}(t) + \mathrm{diag}\{b\} \sin(w t),
\end{equation}
Where $b$ and $w$ are vectors with the same dimensionality as $\theta$, $\mathrm{diag}\{\}$ is the operator that constructs a diagonal matrix from a vector, and $\sin$ is applied element-wise. In all the experiments, $b_{1,2,3} = [0.2, 0.2, 0.5]^T$ millimeters, $b_{4,5,6} = [0.675, 0.675, 0.675]^T$ degrees
and $w = [0.9, 0.83, 0.7, 1.05, 1.0, 0.95]^T$ Hz. \blue{These perturbation parameters were chosen by hand, without much difficulty. It is likely that more performant parameters exist.}

Following the extremum seeking control pipeline, illustrated by Figure \ref{fig:system}, the objective $L$ is sampled at each $\theta$, resulting in a signal $L(t)$. $L(t)$ is high pass filtered with a first-order high pass filter with a $0.7$ Hz cutoff, demodulated with $\sin(wt)$, and low pass filtered with a first-order filter with a cutoff of $1.59$ Hz. This final signal is the feedback error multiplied by a control gain $k$ to determine $\dot{\hat{\theta}}$. Precisely,
\begin{equation}
\dot{\hat{\theta}}(t) = -\mathrm{diag}\{k\} \, \Big[g_{lpf} * x_{demod} \, \big[g_{hpf} * L\big]\Big](t).
\end{equation}
Here $x_{demod}(t) = \mathrm{diag}\{\sin(wt)\}$ is the demodulating signal, $g_{lpf}$ and $g_{hpf}$ are the impulse responses of the low and high pass filters respectively, $*$ denotes convolution, $k = [0.7, 1.1, 0.7, 10.0, 10.0, 10.0]^T$.
A UR10 robot arm tracked $\theta$ using linear servoing.

\begin{figure}[t]
  \centering
  \vspace*{0.1in}
  \includegraphics[width=1.0\linewidth]{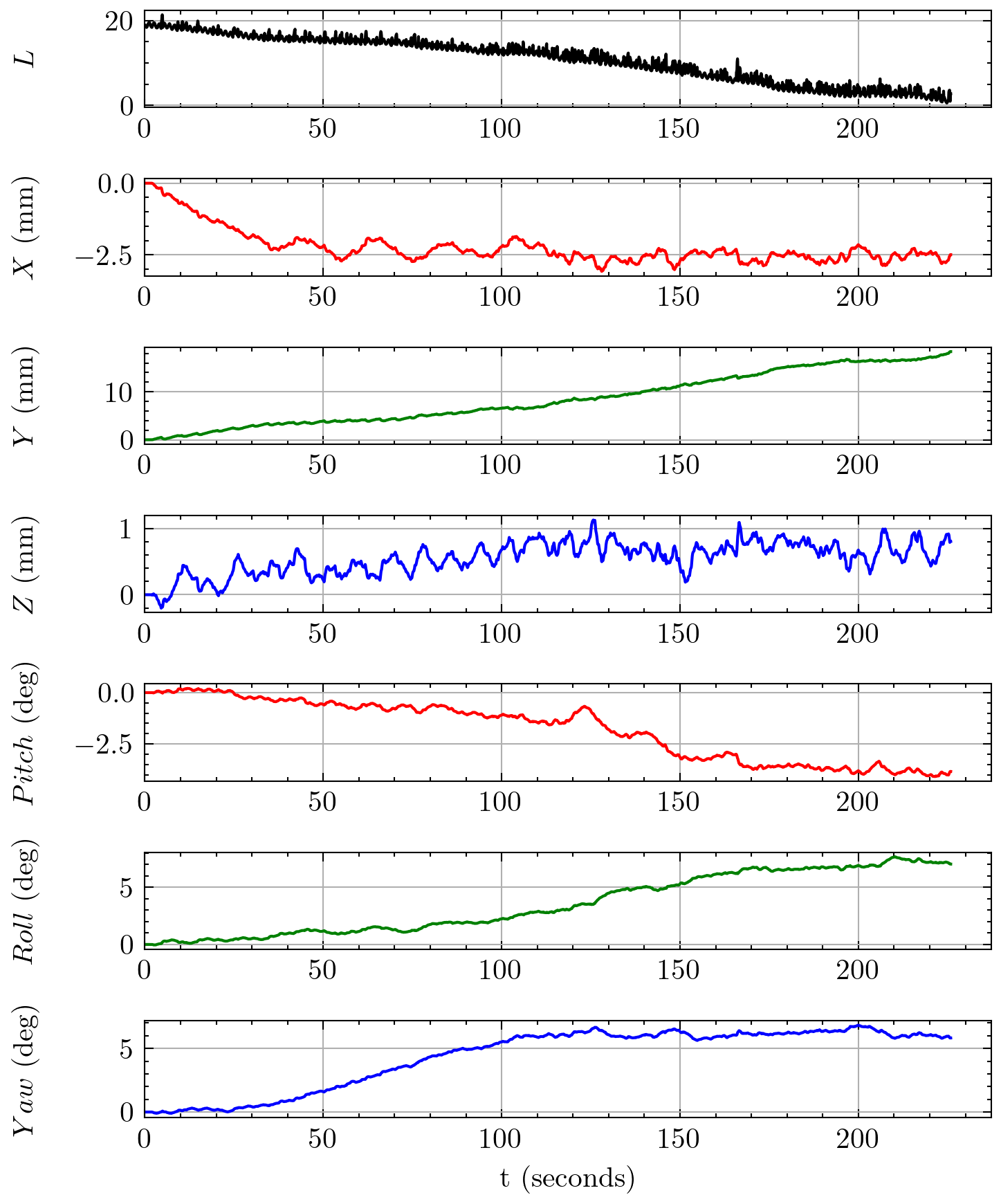}
  \caption{The loss and estimated parameters (end-effector pose) converge as the key is inserted. The Y parameter increases steadily because it corresponds to the insertion axis. The trial pictured was initialized with $[1.1, 0.0]$ millimeters of translation along the X, Z axis and $[3.4, -7.4, 5.7]$ degrees of rotation about the X, Y, Z axis.
  }
  \label{fig:controlsignals_success}
\end{figure}

\section{EXPERIMENTS AND RESULTS}

\blue{Three} sets of experiments were performed involving four distinct lock types \blue{and five objects from the AutoMate assembly benchmark \cite{tang2024automate}}, \mgn{representing complex and basic geometry respectively}.
The first experiment involved 240 trials, and the second 120, \blue{and the third 90,} for a total of \blue{450} trials. The first experiment tested perturbing \mgn{a single degree of freedom (DOF)} from an initial pose aligned with the lock entrance. The results demonstrate which types of perturbations the method is sensitive to. The second experiment tests \mgn{initialization from randomly sampled 6-DOF poses}, thus testing the methods robustness to significant misalignment. The third experiment tests the assumption of initial alignment with the hole, up to a tolerance, by using a camera to estimate the pose of the insertion target and subsequently set the initial position. \mgn{Figure \ref{fig:controlsignals_success} shows the objective and pose during a key in lock insertion starting from a random 6-DOF pose.}

\blue{We establish a baseline for the extremum seeking control law by comparing it to CMA-ES, a gradient free optimization algorithm that is widely used. \mgn{Instead of wiggling, it randomly samples}. Additionally, in the vision initialized trials, we test on the same five objects used in AutoMate's \cite{tang2024automate} vision-initialized trials. AutoMate's baselines are trained on the objects tested on. To our method the objects were unseen.}

\red{Other approaches based on RL were not replicated due to missing code \cite{dong2021tactilerlinsertiongeneralizationobjects, zhang2024multiplepeginholeassemblytightly, yan_2021learning, kamijo2023tactilebasedactiveinferenceforcecontrolled, guo2025srsaskillretrievaladaptation, marougkas2025integratingmodelbasedcontrolrl, li2025easyinsertdataefficientgeneralizableinsertion}, unfeasible assumptions on the objects \cite{kim2022activeextrinsiccontactsensing, dong2019tactile, Gubbi_2020, Gibbons_2023, chen2025robustpeginholeassemblyuncertainties}, starting error limited to certain axis \cite{yan_2021learning}, prior knowledge of contact model \cite{Park_2017}, perfect knowledge of desired final pose \cite{wu2024tacdiffusion},  prior knowledge of the object model \brn{during insertion}\cite{kamijo2023tactilebasedactiveinferenceforcecontrolled}, \mgn{or incompatible sensors and contemporaneity \cite{zhao2025touchbeginsvisionends}}.
}

\begin{figure}[t]
  \centering
  \includegraphics[width=1.0\linewidth]{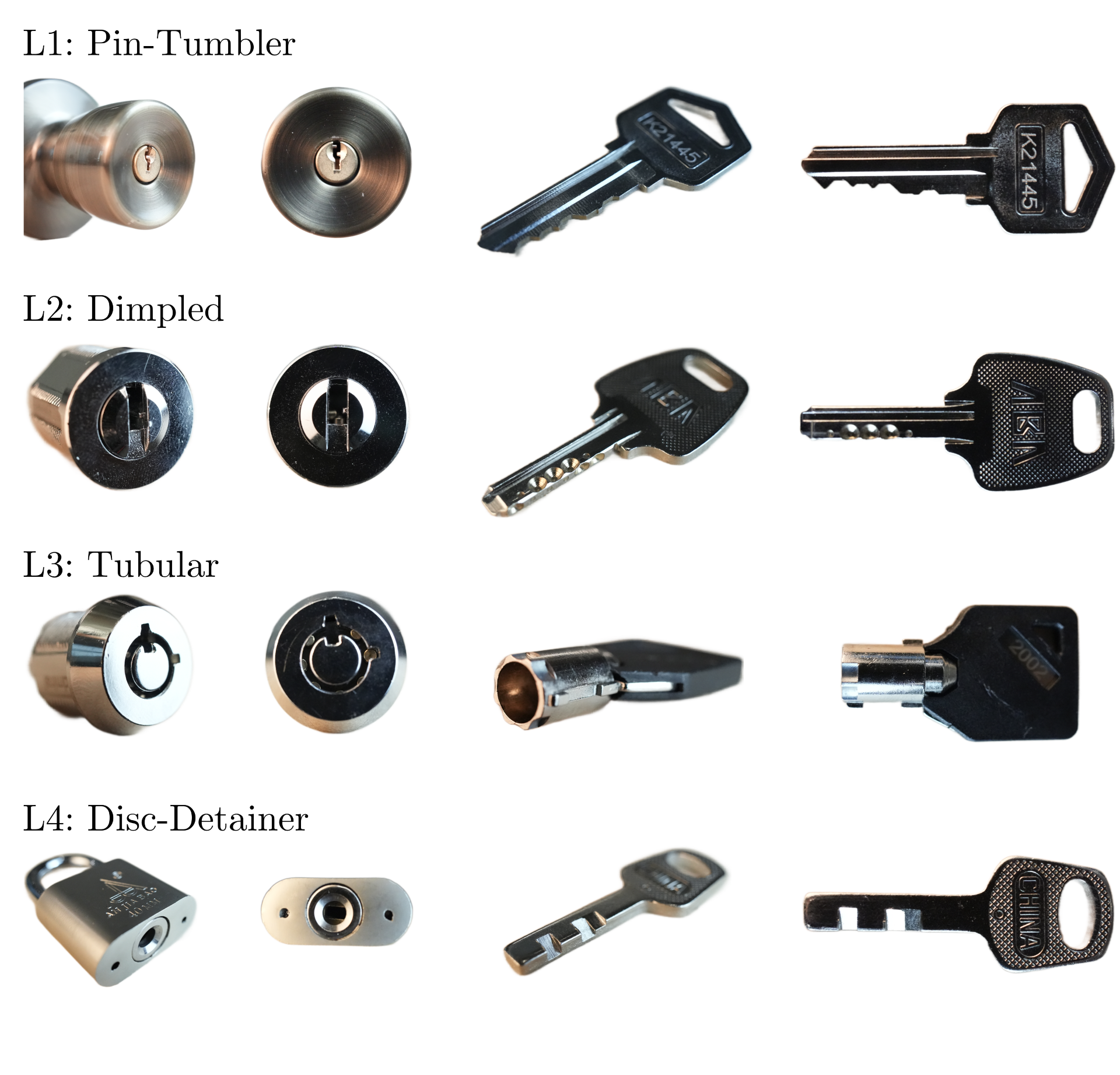}
  \caption{The four types of key and lock pairs tested. L1 is a common pin-tumbler lock used on front doors. L2 is a dimpled cam lock that uses pin-tumblers like L1, but they also press on the sides of the key into the dimples. L3 is a tubular lock with a circular shape that must be pressed into the lock's circular opening. L4 is a disc-detainer which features rotating discs inside the lock that must be aligned.}
  \label{fig:locks}
\end{figure}

\subsection{Lock Types}

\begin{table*}[h]
\vspace{7pt}
\centering
\setlength{\tabcolsep}{3pt}
\resizebox{\textwidth}{!}{
\begin{tabular}{c cccc|cccc|cccc|cccc|cccc || cccc|cccc}
\hline
& \multicolumn{8}{c|}{\textbf{Translation Misalignment (mm)}} & \multicolumn{12}{c||}{\textbf{Rotation Misalignment (deg)}} & \multicolumn{8}{c}{\textbf{CMA Rotation Misalignment (deg)}} \\
\cline{2-29}
\textbf{Lock} & \multicolumn{4}{c}{\textbf{X}} & \multicolumn{4}{c|}{\textbf{Z}} & \multicolumn{4}{c}{\textbf{X}} & \multicolumn{4}{c}{\textbf{Y}} & \multicolumn{4}{c||}{\textbf{Z}} & \multicolumn{4}{c}{\textbf{X}} & \multicolumn{4}{c}{\textbf{Z}} \\
\cline{2-29}
& \textbf{\scriptsize -2.5} & \textbf{\scriptsize -1.9} & \textbf{\scriptsize 1.9} & \textbf{\scriptsize 2.5} & \textbf{\scriptsize -2.5 } & \textbf{\scriptsize -1.9 } & \textbf{\scriptsize 1.9 } & \textbf{\scriptsize 2.5 } & \textbf{\scriptsize -10 } & \textbf{\scriptsize -5 } & \textbf{\scriptsize 5 } & \textbf{\scriptsize 10 } & \textbf{\scriptsize -10 } & \textbf{\scriptsize -5 } & \textbf{\scriptsize 5 } & \textbf{\scriptsize 10 } & \textbf{\scriptsize -10 } & \textbf{\scriptsize -5 } & \textbf{\scriptsize 5 } & \textbf{\scriptsize 10 } & \textbf{\scriptsize -5 } & \textbf{\scriptsize -2 } & \textbf{\scriptsize 2 } & \textbf{\scriptsize 5 } & \textbf{\scriptsize -5 } & \textbf{\scriptsize -2 } & \textbf{\scriptsize 2 } & \textbf{\scriptsize 5} \\
\hline
\multicolumn{29}{c}{\textbf{Success Rate (\%)}} \\
\hline
\textbf{L1} & 100 & 100 & 100 & 100 & 100 & 100 & 100 & 0 & 100 & 100 & 100 & 100 & 100 & 100 & 100 & 33 & 100 & 100 & 100 & 100 & 0 & 100 & 67 & 0 & 0 & 33 & 0 & 0 \\
\textbf{L2} & 67 & 67 & 100 & 0 & 33 & 100 & 33 & 67 & 100 & 100 & 100 & 67 & 100 & 100 & 100 & 100 & 67 & 100 & 67 & 67 & 0 & 0 & 33 & 0 & 0 & 0 & 33 & 0 \\
\textbf{L3} & 100 & 100 & 100 & 0 & 0 & 100 & 100 & 0 & 100 & 67 & 100 & 100 & 0 & 100 & 33 & 0 & 100 & 33 & 67 & 100 & 33 & 100 & 100 & 67 & 100 & 67 & 67 & 67 \\
\textbf{L4} & 100 & 100 & 100 & 100 & 100 & 100 & 100 & 100 & 100 & 100 & 100 & 67 & 100 & 100 & 100 & 100 & 100 & 100 & 100 & 100 & 67 & 100 & 100 & 33 & 67 & 0 & 100 & 0 \\
\hline
\multicolumn{29}{c}{\textbf{Insertion Time (sec)}} \\
\hline
\textbf{L1} & 322 & 96 & 93 & 117 & 215 & 106 & 93 & - & 232 & 116 & 113 & 217 & 505 & 91 & 85 & 120 & 160 & 123 & 136 & 181 & - & 249 & 265 & - & - & 231 & - & - \\
\textbf{L2} & 335 & 195 & 164 & - & 256 & 196 & 739 & 338 & 188 & 153 & 153 & 185 & 201 & 308 & 147 & 203 & 300 & 164 & 170 & 490 & - & - & 213 & - & - & - & 267 & - \\
\textbf{L3} & 335 & 89 & 83 & - & - & 118 & 53 & - & 145 & 36 & 80 & 155 & - & 44 & 62 & - & 85 & 41 & 52 & 106 & 73 & 48 & 65 & 56 & 61 & 55 & 47 & 112 \\
\textbf{L4} & 86 & 90 & 97 & 88 & 104 & 95 & 106 & 86 & 100 & 100 & 104 & 99 & 111 & 91 & 90 & 68 & 115 & 113 & 108 & 96 & 142 & 162 & 208 & 185 & 220 & - & 194 & - \\
\hline
\end{tabular}}
\caption{Success rate and insertion time versus single parameter perturbations of initial pose.}
\label{table:merged_results}
\end{table*}

\mgn{The} four locks tested \mgn{are} illustrated in Figure \ref{fig:locks}.  \mgn{All feature complex geometry.} The first lock is a cylindrical pin-tumbler lock.
The ridges on the top of the key actuate the pin tumblers inside the lock. The second is a dimpled key cam lock that uses dimples on the sides of the key to actuate pin-tumblers inside the lock. Compared to a standard pin-tumbler, dimpled keys require some variation in force to insert due to the way the pin-tumblers press into the dimples. The third lock is a tubular cam lock. As shown in Figure \ref{fig:locks} the tubular key is circular instead of rectangular and thus has a significantly different form than the pin-tumbler and dimpled keys.
Finally, the fourth lock is a padlock disk-detainer. Disc-detainer locks use rotating disks that must be aligned.
\blue{The locks also have differently shaped front faces. L1 and L4's \mgn{have front faces that generally slope toward the opening.} However, L2 and L3 feature regions near the opening that slope away, or are flat.}

\subsection{Initial Pose Perturbed on One Axis}

The first experiment demonstrates the robustness to perturbations along one translation or rotational axis of the initial key pose away from alignment with the lock's keyhole. For each lock, translation perturbations of -2.5, -1.9, 1.9, 
and 2.5 millimeters were tested along the plane parallel to the lock surface. Additionally, initial rotation perturbations of -10, -5, 5, and 10 degrees were applied to each orientation axis individually. A 0-millimeter and 0-degree setting corresponds to a near-perfect alignment between the key and the lock. \red{The translation misalignment considered is larger than the hole size. In contrast, state-of-the-art works that, like ours, do not incorporate a search strategy \cite{kim2022activeextrinsiccontactsensing, Park_2020b}, assume smaller translations relative to hole size.} Three tests were performed per initial pose.
The results are given in Table \ref{table:merged_results}.

The success rate over the smaller perturbations (-1.9 and 1.9 mm in translation) and (-5 and 5 degrees in rotation)  was 90.8\% over 120 trials with a mean insertion time of 122 seconds. The success rate over the large perturbations (-2.5 and 2.5 mm in translation) and (-10 to 10 degrees in rotation) dropped to 76.7\% over 120 trials with a mean insertion time of 171 seconds. The overall success rate over the 240 trials was 83.8\% with a mean insertion time of 147 seconds. 

Lock L4 (Disc-Detainer) was the easiest as only one trial failed, resulting in a 98\% success rate overall with mean insertion times of 98 seconds. Lock L1, L2, and L3 achieved 93\%, 77\%, and 65\% overall with mean insertion times of 169,  232, and 105 seconds respectively. The drop in performance on lock L2 is expected because the lock surface \mgn{mainly slopes away from the keyhole,} which makes it difficult for the extremum seeking control to enter. Similarly, lock L3 is flat on the surface of the keyhole and so it proved difficult for the extremum seeking control to find the hole, especially when the perturbations were large.

\blue{As a baseline,}
\red{we conducted experiments using CMA-ES \cite{auger2012tutorial} using \blue{smaller} angular perturbations of -5, -2, 2, and 5 degrees along the X and Z axes. \blue{The overall success rate was 42\% over 96 trials. Detailed} results are reported on the right side of Table \ref{table:merged_results}. CMA-ES fails with small 2-degree perturbations, whereas extremum seeking performs well at 5 and even 10-degree perturbations. CMA-ES was not tested with translation perturbations because the performance on rotation was convincingly poor.}
\blue{CMA-ES had to be modified to avoid breaking the GelSight. Sigma adaptation was disabled as it excessively increased the pose perturbations. Even then, the movements to the requested poses were frequently aborted because the measured strain exceeded a safety threshold.}

\begin{table*}[h]
\vspace{7pt}
\centering
\begin{tabular}{c c@{\hspace{5pt}}c@{\hspace{5pt}}c@{\hspace{5pt}}c|c@{\hspace{5pt}}c@{\hspace{5pt}}c@{\hspace{5pt}}c|c@{\hspace{5pt}}c@{\hspace{5pt}}c@{\hspace{5pt}}c|c@{\hspace{5pt}}c@{\hspace{5pt}}c@{\hspace{5pt}}c|c@{\hspace{5pt}}c@{\hspace{5pt}}c@{\hspace{5pt}}c}
\hline
 & \multicolumn{8}{c}{\textbf{Translation Misalignment (mm)}} & \multicolumn{12}{c}{\textbf{Rotation Misalignment (deg)}} \\
\cline{2-21}
\textbf{Type of Lock} & \multicolumn{4}{c}{\textbf{X}} & \multicolumn{4}{c|}{\textbf{Z}} & \multicolumn{4}{c}{\textbf{X}} & \multicolumn{4}{c}{\textbf{Y}} & \multicolumn{4}{c}{\textbf{Z}} \\ \cline{2-21}
& \tiny \textbf{(-2.5,-1.9)} & \tiny \textbf{(-1.9,0)} & \tiny \textbf{(0,1.9)} & \tiny \textbf{(1.9,2.5)} & \tiny \textbf{(-2.5,-1.9)} & \tiny \textbf{(-1.9,0)} & \tiny \textbf{(0,1.9)} & \tiny \textbf{(1.9,2.5)} & \tiny \textbf{(-10,-5)} & \tiny \textbf{(-5,0)} & \tiny \textbf{(0,5)} & \tiny \textbf{(5,10)} & \tiny \textbf{(-10,-5)} & \tiny \textbf{(-5,0)} & \tiny \textbf{(0,5)} & \tiny \textbf{(5,10)} & \tiny \textbf{(-10,-5)} & \tiny \textbf{(-5,0)} & \tiny \textbf{(0,5)} & \tiny \textbf{(5,10)}\\

\textbf{no. of trials} & \tiny \textbf{2} & \tiny \textbf{9} & \tiny \textbf{15} & \tiny \textbf{4} & \tiny \textbf{2} & \tiny \textbf{10} & \tiny \textbf{13} & \tiny \textbf{5} & \tiny \textbf{8} & \tiny \textbf{8} & \tiny \textbf{9} & \tiny \textbf{5} & \tiny \textbf{6} & \tiny \textbf{6} & \tiny \textbf{11} & \tiny \textbf{7} & \tiny \textbf{10} & \tiny \textbf{7} & \tiny \textbf{6} & \tiny \textbf{7}\\
\hline
\multicolumn{21}{c}{\textbf{Success Rate (\%)}} \\
\hline

\textbf{L1} & 100 & 67 & 73 & 100 & 0 & 90 & 85 & 60 & 100 & 62 & 78 & 60 & 67 & 67 & 91 & 71 & 100 & 100 & 67 & 29 \\ 
\textbf{L2} & 50 & 67 & 80 & 0 & 100 & 80 & 54 & 40 & 62 & 62 & 78 & 40 & 67 & 67 & 64 & 57 & 70 & 43 & 67 & 71\\
\textbf{L3} & 50 & 56 & 47 & 0 & 0 & 50 & 46 & 40 & 62 & 12 & 33 & 80 & 33 & 67 & 36 & 43 & 70 & 14 & 67 & 14\\
\textbf{L4} & 100 & 100 & 100 & 100 & 100 & 100 & 100 & 100 & 100 & 100 & 100 & 100 & 100 & 100 & 100 & 100 & 100 & 100 & 100 & 100\\ \hline
\multicolumn{21}{c}{\textbf{Insertion Time (sec)}} \\
\hline

\textbf{L1} & 125 & 277 & 204 & 169 & - & 226 & 177 & 285 & 205 & 261 & 173 & 225 & 173 & 161 & 270 & 449 & 151 & 161 & 167 & 383 \\ 
\textbf{L2} & 1360 & 387 & 429 & - & 545 & 416 & 385 & 860 & 389 & 430 & 594 & 289 & 303 & 571 & 507 & 594 & 346 & 556 & 410 & 587\\
\textbf{L3} & 210 & 301 & 406 & - & - & 325 & 341 & 445 & 277 & 342 & 403 & 406 & 406 & 250 & 324 & 170 & 487 & 320 & 262 & 419\\
\textbf{L4} & 135 & 130 & 138 & 125 & 132 & 107 & 139 & 173 & 114 & 109 & 123 & 223 & 165 & 103 & 150 & 106 & 106 & 146 & 129 & 153\\ \hline
\end{tabular}
\caption{\red{Marginal distributions of success and insertion time for Extremum Seeking Control with \blue{30} random initial poses tested on all four locks. Our method achieves an average insertion time of 262 seconds with a 71\% success rate. The baseline, CMA-ES, \blue{failed on 5} consecutive trials on L1, L2, and L4 and succeeded on 2 out of 5 trials on L3.}}
\label{table:accuracyrandom}
\end{table*}

\begin{table*}[t]
\centering
\scriptsize
\resizebox{\textwidth}{!}{
\begin{tabular}{l|cccc|ccccc}
\hline
{} &    \textbf{L1} &        \textbf{L2} &  \textbf{L3} &        \textbf{L4} &           \textbf{00340} &          \textbf{00320} &          \textbf{00346} &            \textbf{00015} &           \textbf{00296} \\
\hline

\textbf{Pitch} (deg) &  0.504 ± 0.415 &  0.522 ± 0.463 &  2.174 ± 0.710 &  0.636 ± 0.521 &   0.882 ± 0.545 &  4.677 ± 2.174 &  4.677 ± 2.174 &    3.064 ± 1.971 &   1.768 ± 0.913 \\
\textbf{Roll} (deg)  &  2.490 ± 0.681 &  1.337 ± 0.500 &  0.879 ± 0.325 &  0.511 ± 0.319 &  96.289 ± 0.470 &  2.640 ± 0.646 &  2.640 ± 0.646 &  62.755 ± 35.238 &  89.953 ± 0.922 \\
\textbf{Yaw} (deg)   &  2.298 ± 0.577 &  1.441 ± 0.850 &  0.696 ± 0.555 &  0.922 ± 0.631 &   0.849 ± 0.660 &  3.319 ± 2.169 &  3.319 ± 2.169 &    1.984 ± 1.293 &   2.547 ± 0.913 \\
\hline

\textbf{X} (m)       &  0.002 ± 0.001 &  0.001 ± 0.001 &  0.001 ± 0.001 &  0.001 ± 0.001 &   0.002 ± 0.001 &  0.002 ± 0.002 &  0.002 ± 0.002 &    0.002 ± 0.001 &   0.001 ± 0.001 \\
\textbf{Y} (m)       &  0.002 ± 0.001 &  0.003 ± 0.001 &  0.003 ± 0.001 &  0.003 ± 0.001 &   0.002 ± 0.001 &  0.002 ± 0.001 &  0.002 ± 0.001 &    0.003 ± 0.002 &   0.002 ± 0.001 \\
\textbf{Z} (m)       &  0.002 ± 0.001 &  0.003 ± 0.002 &  0.005 ± 0.002 &  0.003 ± 0.002 &   0.004 ± 0.002 &  0.007 ± 0.002 &  0.007 ± 0.002 &    0.001 ± 0.001 &   0.004 ± 0.002 \\
\hline
\end{tabular}}
\caption{\mgn{Vision based initialization's rotation and translation error (mean ± std) estimated from 100 samples per object. Objects with high roll error are symmetric about the roll axis.}}
\label{tab:vision_accuracy}
\end{table*}

\subsection{Initial Pose Determined Randomly}

Robustness to initial pose was evaluated using 30 random samples of end-effector pose. Translations parallel to the lock surface were drawn from a uniform distribution over -2.5 to 2.5 mm, and rotations in Euler angles were drawn from a uniform distribution over -10 to 10 degrees. The distributions of the success rates are given in Table \ref{table:accuracyrandom}.

Our method's average success rate over the uniformly sampled smaller translation perturbations (-1.9 to 1.9 mm) was 74\% in X and 75\% in Z with mean insertion times of 269 and 240 seconds respectively. Over the smaller rotational perturbations (-5 to 5 degrees), the expected success rate was 66\% in X, 73\% in Y (insertion direction), and 69\% in Z with mean insertion times of 274, 241, and 253 seconds respectively. The success rate over larger translation perturbations (-2.5 to 2.5 mm) was 58\% in X and 57\% in Z with mean insertion times of 233 and 355 seconds respectively. Over the larger rotational perturbations (-10 to 10 degrees), the expected success rate was 77\% in X, 67\% in Y (insertion direction), and 72\% in Z with mean insertion times of 247, 291, and 268 seconds respectively. The overall success rate was 71\% over 120 trials with a mean insertion time of 262 seconds. Lock L4 (Disc-Detainer) was the easiest, with zero failures and a mean insertion time of 134 seconds. Lock L1, L2, and L3 achieved 77\%, 63\%, and 44\% overall with mean insertion times of 210, 465, 351 seconds respectively. \blue{As a baseline, we tested CMA-ES with 5 random poses on each lock and succeeded only twice, both times on L3}.

\subsection{\blue{Vision Based Initialization}}\label{subsec:vision}

\begin{figure}[t]
  \centering
  \includegraphics[width=0.9\linewidth]{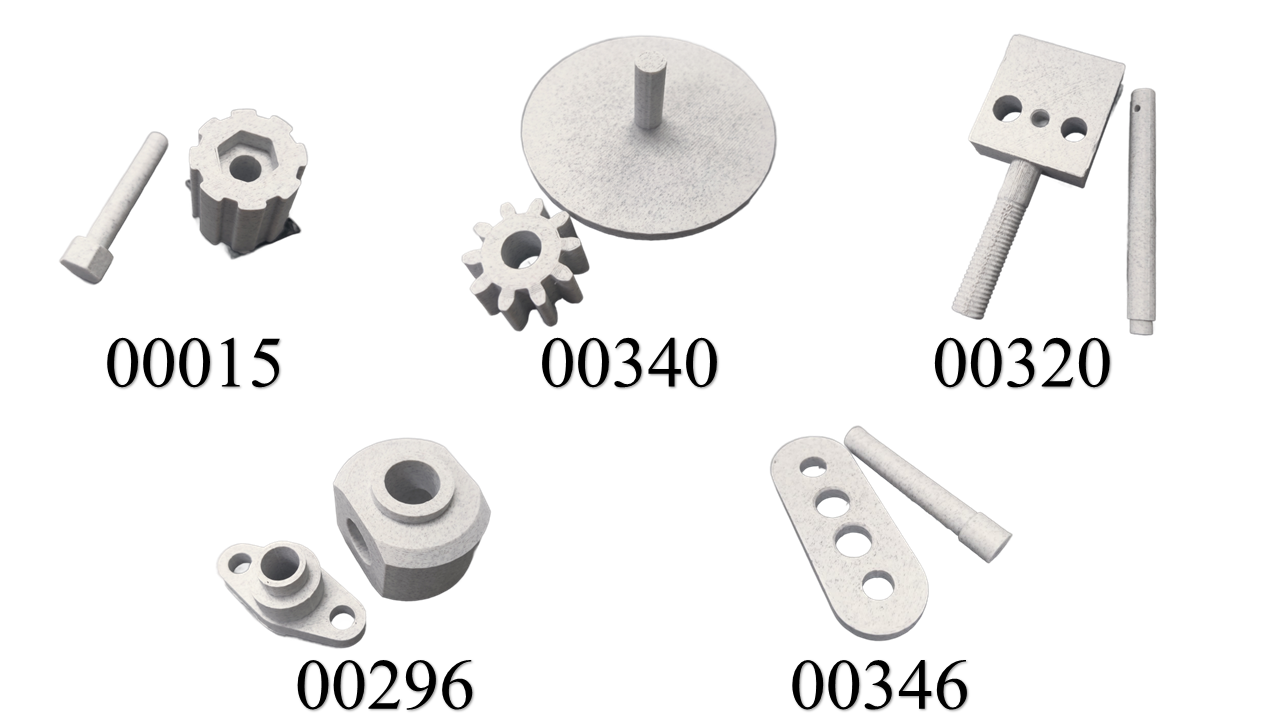}
  \caption{\mgn{We evaluate vision initialized insertion on 5 objects with basic geometry from a robot assembly benchmark \cite{tang2024automate}.}}
  \label{fig:automate_objects}
\end{figure}

\blue{The pose uncertainties assumed by the previous experiments were validated with two experiments using vision based initialization. Similar to AutoMate, a hand mounted camera was used to estimate the object's pose using \cite{wen2024foundationpose} and \cite{ravi2024sam}. Experiments were conducted using the 4 locks and \mgn{also 5 assemblies from a robot assembly benchmark \cite{tang2024automate}.} The assemblies are pictured in Figure \ref{fig:automate_objects}. One hundred random poses sampled from a cube 20 cm on each side, with the closest face 20 cm from the opening. \mgn{Orientation was sampled uniformly from $SO(3)$ and rejected if the plug was outside the field-of-view or the roll was greater than 45 degrees. The resulting yaw, pitch, and roll fell between approximately $\pm$40,  $\pm$25, and $\pm$45 degrees respectively. The pose was then estimated from vision and the resulting} translation and orientation error and standard deviation for each object are listed in Table \ref{tab:vision_accuracy}. \mgn{The errors are typically within a few millimeters or degrees, which is well within the margins tolerated by our method on locks.}}

\blue{
Additionally, 10 vision initialized insertion trials were conducted on each object (90 trials total). \mgn{The starting pose was randomly sampled as when estimating pose uncertainty.} The success rates are reported in Table \ref{tab:insertion_success_vision}. The average success rate for the locks and assemblies was 75\% and 98\% respectively. The performance on locks was comparable to the average performance in the previous experiments, validating our assumptions. The average performance on AutoMate's assemblies was higher than AutoMate's vision-initialized generalist and specialist baseline's which achieved 86\% and 90\% respectively.
It should be noted that AutoMate's vision-initialized trials also estimated the pose of the object in hand. However, our average vision-initialized performance is higher than AutoMate's highest manually initialized basline performance, which was 96\%. These results clearly show that our method generalizes well to other types of insertion, significantly outperforming the state-of-the-art approach without per-object training, or in general, learning.}

\mgn{AutoMate's baseline was evaluated on assemblies with 1 millimeter, 45 degree chamfers on  contacting edges. The released dataset is not chamfered. For our evaluation, we applied smaller 0.707 millimeter chamfers. We also evaluated on non-chamfered assemblies, resulting in a substantially lower average score of 50\% (Ours-NC in Table \ref{tab:insertion_success_vision}). These results indicate our method benefits from a small amount of chamfering. It is unknown how AutoMate performs on non-chamfered assemblies. While AutoMate's training code is available, the code for deploying on a real robot is not.}

\begin{table}[t]
\centering
\scriptsize
\resizebox{\columnwidth}{!}{
\begin{tabular}{l|c||l|c|c|c}
\hline
\textbf{Lock} & \textbf{Ours} & \textbf{Assembly} &\textbf{AutoMate \cite{tang2024automate}} & \textbf{Ours} & \textbf{Ours-NC}  \\
\hline
L1      & 80  & 00340    & 80 & 100 & 30 \\
L2      & 80  & 00320    & 80 & 100 & 80 \\
L3      & 50  & 00346    & 80 & 100 & 40 \\
L4      & 90  & 00015    & 100 & 100 & 60 \\
        &     & 00296    & 90 & 90 & 40 \\
\hline
\textbf{Mean} & \textbf{75} & \textbf{Mean} & \textbf{86} & \textbf{98} & \textbf{50}\\
\hline
\end{tabular}}
\caption{\mgn{Vision initialized success rate (\%) on locks and assemblies (10 trials each). AutoMate's baseline is trained on the assemblies.
AutoMate evaluated on chamfered assemblies, but the released dataset is not chamfered. Ours-NC denotes our method on non-chamfered assemblies.}}
\label{tab:insertion_success_vision}
\end{table}

\section{DISCUSSION AND FUTURE WORK} 
\mgn{The value of a closed loop wiggling process guided by tactile feedback in complex insertion problems has been demonstrated using an extremum seeking controller. A single set of a parameters generalizes to multiple locks, which feature complex geometry, and achieves remarkable success on a robot assembly benchmark which features basic geometry.}

\red{The method significantly outperforms a baseline based on CMA-ES which struggles to correct the orientation of the key so that it can enter the keyhole. The difference in performance can be explained by considering the major difference in sampling strategy between the two algorithms. Extremum seeking continuously moves the key into different poses with a wiggling like motion while simultaneously sampling the sensor. This allows it to press the key into the lock while \mgn{also} adjusting orientation. In contrast, CMA-ES samples \mgn{random} poses which results in a much slower sampling rate and \mgn{far fewer orientation samples while pressing into the lock.}}

\mgn{The performance on the assembly benchmark was significantly higher than that of the learning-based approach, validating the importance of closed-loop wiggling.
Moreover, these experiments validate the effectiveness of the proposed approach across a broader range of geometries and insertion types. For instance, objects 00340 and 00296 represent cases of ``inverse'' insertion, where the socket is held in hand instead of the plug. Finally, the results reinforce the assumption that key insertion constitutes a particularly challenging task.}

\mgn{Next, we analyze the primary reasons for failure on each lock}.
Lock L1 (pin-tumbler) has sharp edges along the top and bottom of the key hole that the key can become stuck in. In this case, our method is too naive to detect that the key is stuck, or wedged, and does not back away (12 failure cases). Lock L2 (dimpled) has sharp edges on the key that sometimes became caught on the entrance of the lock (7 failure cases). Additionally, L2 has a raised keyhole as can be seen in Figure \ref{fig:locks}. Thus, the extremum seeking control law sometimes moved the key over the edge and got stuck (11 failure cases). Finally, lock L3 (tubular) achieved the lowest success rate because orientation about the Y axis (insertion axis) is much more important than in L1, L2, or L4. Additionally, the surface of the lock is flat, and thus provides little to no information for the extremum seeking control law to use to proceed into the lock (33 failure cases). These two issues also occasionally caused a maximum limit on strain to be exceeded (3 failure cases). The reasons for five cases of failure were not recorded.

\mgn{Most failure modes} are due to 
the direct application of extremum seeking control. \mgn{While extremum seeking can use its sinusoidal perturbations to escape small local minima and saddle points, such as those caused by the the teeth of a key, it cannot always escape.} 
\mgn{Regardless, the performance on the key and lock problem, which features complex geometry, is striking.}
\mgn{T}here are many ways to improve. \mgn{In particular, approaches that incorporate learning in a manner similar to \cite{tang2024automate} and \cite{wu2024tacdiffusion} are promising. Additionally, it is of interest to learn task specific wiggling patterns and consider more specific tactile feedback than the proposed displacement based strain analogue. Further, we note that the primary technical assumption of our implementation, grasping planar and rigid local surfaces whose displacement can be represented as a homography, could be removed by replacing the tactile strain estimator, without altering the overall approach.} \mgn{Finally, the current work relies on a tight grasp; future work can consider adjusting grasp strength and compensating for slippage}.

The current approach also admits several immediate paths towards improving the mean insertion time and success rate. In particular, the control parameters were tuned only to demonstrate basic efficacy
Additionally, the objective function is currently linear in each argument and \mgn{which results in} the slow, linear convergence seen in Figure \ref{fig:controlsignals_success}.

\begin{table}[H]
\small
  \centering
  \begin{tabular}{@{}lp{3.5cm}p{2.5cm}}
    %\toprule
    \\[0.03cm]
 \textbf{Type} & \textbf{Examples} & \textbf{Tactile Modalities} \\
    \midrule
    Direct &
    \parbox{3.5cm}{Key, Seatbelt} &
    \parbox{3.5cm}{Grasp strain}\\
    \midrule
    Through &
    \parbox{3.5cm}{Cable routing \\ Putting on belt} &
    \parbox{3.5cm}{Grasp strain\\Bimanual contact} \\
    \midrule
    Crowded  &
    \parbox{3.5cm}{Putting a cup in a cupboard \\ Packing a box } &
    \parbox{3.5cm}{Grasp strain \\ Manipulator contact} \\
    \bottomrule
  \end{tabular}
  \caption{\blue{Tasks completed by humans with wiggling that are promising directions for future work.}
  }
  \label{tab:insertionmodes}
\end{table}

\section{CONCLUSION}% AND FUTURE WORK}

\mgn{Closed-loop wiggling using tactile feedback has significant potential in robotic insertion problems involving complex geometries.
Future work can consider many adaptions.}
\mgn{In particular, we propose several insertion problems in Table \ref{tab:insertionmodes} which feature complex geometries that humans often tackle using wiggling. We categorize them as ``direct'', ``through'', and and ``crowded'' and propose the tactile modality that is likely to be most important during the wiggling process.}

% A conclusion section is not required. Although a conclusion may review the main points of the paper, do not replicate the abstract as the conclusion. A conclusion might elaborate on the importance of the work or suggest applications and extensions. 


%%%%%%%%%%%%%%%%%%%%%%%%%%%%%%%%%%%%%%%%%%%%%%%%%%%%%%%%%%%%%%%%%%%%%%%%%%%%%%%%

%%%%%%%%%%%%%%%%%%%%%%%%%%%%%%%%%%%%%%%%%%%%%%%%%%%%%%%%%%%%%%%%%%%%%%%%%%%%%%%%



%\newpage
\section*{Acknowledgments}
The authors thank Antonio Gambale for helpful discussions about lock types and characteristics. The support of the NSF under award OISE 2020624 is gratefully acknowledged. G. M. C. and L. N. acknowledge financial support from the PNRR MUR project PE0000013-FAIR.

\bibliographystyle{IEEEtran}
\bibliography{bibliography}

% \newpage

 

% \end{IEEEbiography}



% \vfill

\end{document}